\documentclass{article}

\usepackage{arxiv}

\usepackage[utf8]{inputenc} 
\usepackage[T1]{fontenc}    
\usepackage{hyperref}       
\usepackage{url}            
\usepackage{booktabs}       
\usepackage{amsfonts}       
\usepackage{nicefrac}       
\usepackage{microtype}      
\usepackage{lipsum}		
\usepackage{graphicx}
\usepackage{natbib}
\usepackage{doi}
\usepackage{multirow}
\usepackage[table]{xcolor}
\definecolor{mygray}{gray}{.9}
\definecolor{sota_blue}{HTML}{0071bc}
\usepackage{bbm}
\usepackage{tikz}
\usepackage{pgfplots}
\usepackage{pgfplotstable}
\usepackage{float}
\usepackage{dblfloatfix}
\usepackage{adjustbox}
\usepackage{makecell}
\usepackage{bm}

\usepackage{algorithmic}
\usepackage{algorithm}
\usepackage{bm}
\usepackage{booktabs}
\usepackage[table]{xcolor}

\usepackage{pgfplots}
\pgfplotsset{compat=newest}
\usepackage{amsmath}
\usepackage{amsthm}
\usepackage{amsfonts}
\usepackage{subcaption}
\usepackage{cleveref}
\usepackage{amssymb}

\def\dabdetr{DAB DETR}

\newcommand\ie{\textit{i.e.}}

\title{Knowledge Distillation via Query Selection for Detection Transformer}

\date{} 					

\author{
{Yi Liu}\\
	Beihang University\\
 \texttt{18373214@buaa.edu.cn} \\
	\And
{Luting Wang} \\
Beihang University\\
\texttt{wangluting@buaa.edu.cn} \\
 \And
 {Zongheng Tang} \\
Beihang University\\
\texttt{tzhhhh123@buaa.edu.cn} \\
 \And
{Yue Liao}\thanks{Corresponding author} \\
Beihang University\\
\texttt{liaoyue.ai@gmail.com} \\
 \And
{Yifan Sun} \\
Baidu Inc\\
\texttt{sunyf15@tsinghua.org.cn} \\
 \And
 {Lijun Zhang} \\
Beihang University\\
\texttt{ljzhang@buaa.edu.cn} \\
 \And
  {Si Liu} \\
Beihang University\\
\texttt{liusi@buaa.edu.cn} \\
 \AND
}
\begin{document}
\maketitle
\begin{abstract}
Transformers have revolutionized the landscape of object detection through the introduction of Detection Transformers (DETRs), acclaimed for their simplicity and efficacy. Despite their advantages, the substantial size of these models poses significant challenges for practical deployment, particularly in resource-constrained environments. This paper addresses the challenge of compressing transformer-based detectors by leveraging knowledge distillation, a technique that holds promise for maintaining model performance while reducing size.
A critical aspect of DETRs' performance is their reliance on queries to interpret object representations accurately. Traditional distillation methods often focus exclusively on positive queries, identified through bipartite matching, neglecting the rich information present in hard-negative queries. Our visual analysis indicates that hard-negative queries, with their focus on foreground elements, are crucial for enhancing distillation outcomes. 
To this end, we introduce a novel Group Query Selection strategy, which diverges from traditional query selection in DETR distillation by segmenting queries based on their Generalized Intersection over Union (GIoU) with ground truth objects, thereby uncovering valuable hard-negative queries for distillation. 
Furthermore, we present the Knowledge Distillation via Query Selection for Detection Transformer (QSKD) framework, which incorporates Attention-Guided Feature Distillation (AGFD) and Local Alignment Prediction Distillation (LAPD). These components optimize the distillation process by focusing on the most informative aspects of the teacher model's intermediate features and output.
Our comprehensive experimental evaluation of the MS-COCO dataset demonstrates the effectiveness of our approach, significantly improving average precision (AP) across various DETR architectures without incurring substantial computational costs. Specifically, the AP of Conditional DETR ResNet-18 increased from $35.8$ to $39.9$, DAB DETR ResNet-18 from $36.2$ to $41.5$, and DINO ResNet-50 from $49.0$ to $51.4$.

\end{abstract}


\section{Introduction}
\label{sec:intro}
Knowledge distillation (KD)~\cite{kd, detection_kd, SONG2024110235}
serves as an effective model compression strategy, enabling the transfer of knowledge from cumbersome teacher models into lightweight student models. 
Its application in object detection has been burgeoning, yet it remains predominantly tailored to CNN-based detectors~\cite{faster_rcnn,fcos, XIE2024110172}
This specificity stems from an architectural mismatch, which renders these methods less effective—or inapplicable—for the nascent transformer-based detectors~\cite{ WANG2023109817, tang2023detr, KORBAN2023109713}.
Given the ascent of transformer-based models to the forefront of object detection, the development of KD algorithms tailored for these architectures is imperative~\cite{detrdistill}.

\begin{figure}[t] 
\begin{center}
    \includegraphics[width=0.8\linewidth]{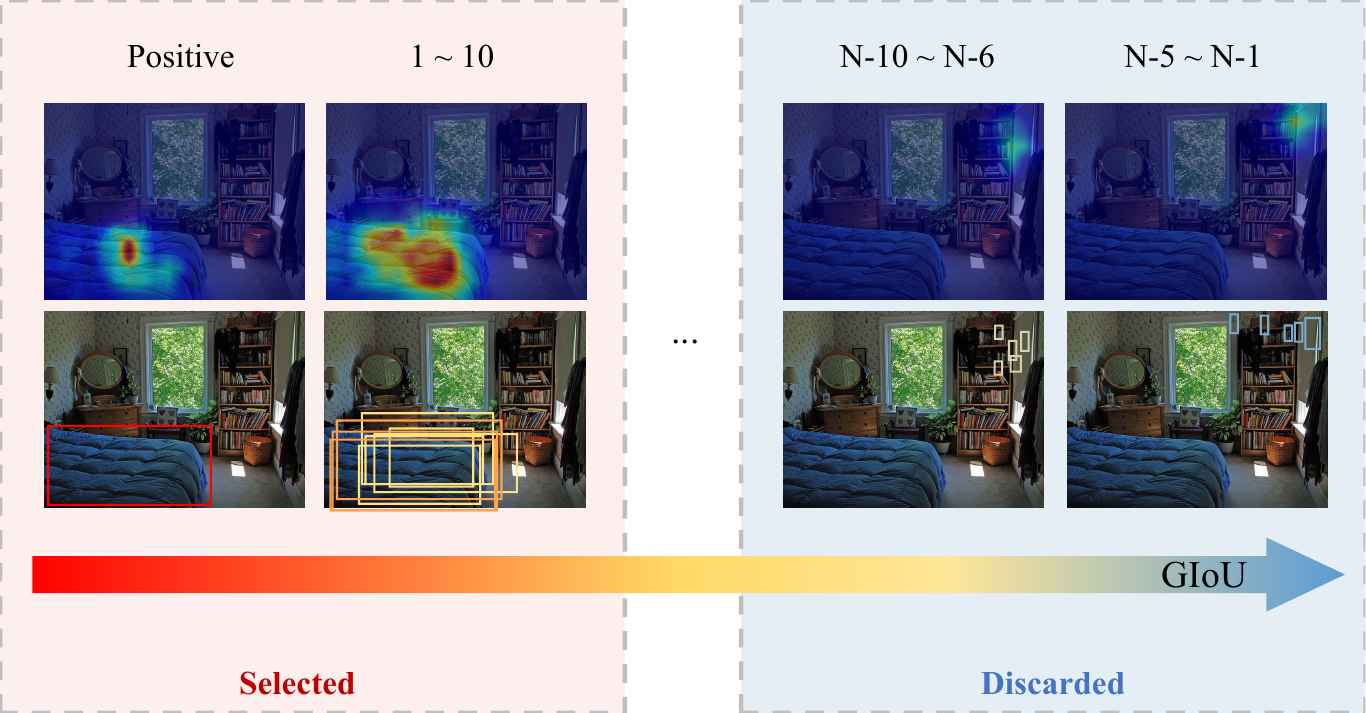}
    \caption{
        Visualization of the attentive regions (the first row) and the predicted boxes (the second row) of N different queries.
        Each of these queries is assigned to the corresponding ground truth object (the bed), selected because their GIoU with the bed is the highest among all objects. The queries are then sorted according to their GIoU metric relative to the bed.
        The first column features the positive prediction from bipartite matching, while the second column shows the top $10$ negative predictions with the highest GIoU metrics.
        The last two columns present the $10$ prediction with the lowest GIoU metrics.
        Based on the hypothesis that queries attending to the foreground regions contain valuable information for distillation, we propose to select the queries with higher GIoU metrics while discarding the ones with lower GIoU metrics.
       }
    \label{fig:queries}
\end{center}
\end{figure}

Object queries are vital to transformer-based detectors due to their ability to predict bounding boxes by gathering relevant information from image features~\cite{detr, meng2021conditional, zhang2022dino, gao2024ease}. 
In this context, object queries are equipped for knowledge extraction from the image features, while simultaneously facilitating the alignment of predictions between teacher and student models~\cite{detrdistill,d3etr}.
Jiahao Chang et al. believe that not all object queries of the teacher should be
equally treated as valuable cues~\cite{detrdistill} for feature distillation, thus solely selecting positive queries based on bipartite matching results.
However, the bipartite matching overlooks many queries valuable for distillation, leading to suboptimal and inefficient distillation processes.
We contend that an effective approach to DETR distillation should concentrate on informative queries, which are not adequately represented by solely positive queries or the aggregate of all queries. 

We argue that hard-negative queries, which also accurately locate objects but fail to match with ground truth in bipartite matching, contain valuable knowledge for distillation.
We first analyzed the average number of queries associated with each ground truth object at various Generalized Intersection over Union (GIoU) thresholds, as detailed in ~\Cref{tab:intro}.
By adjusting the GIoU threshold, we selectively filter the matching query for each object. 
The statistics suggests that various queries, beyond just the positive query, are capable of producing accurate predictions that align with the ground truth object across different GIoU thresholds.
We further visualize the association between object queries and image features in \Cref{fig:queries}.
Though the positive query achieves the maximum GIoU metric, it only focuses on a minor portion of the object area, limiting the scope of knowledge extraction.
In contrast, as the GIoU metric decreases, the queries spread out to the vicinity of the object, encapsulating more knowledge about the object.
We believe that these hard-negative queries with accurate spatial positioning, can lead to enhanced distillation outcomes.
Building on these insights, we develop Group Query Selection which groups queries based on their GIoU with different objects and then selects appropriate queries in each group by GIoU threshold.
Through this process, we effectively identify and select valuable hard-negative queries for distillation. 
\begin{table}[]
\centering
\caption{The average query number assigns to ground truth boxes according to GIoU under different GIoU thresholds. Not only do positive queries provide accurate spatial positioning, but certain hard-negative queries can also precisely pinpoint the locations of objects.}
\label{tab:intro}
\setlength{\extrarowheight}{3pt}{
\setlength{\tabcolsep}{0.9mm}{
\begin{tabular}{@{}c|c|c|c|c|c@{}}
\toprule
GIoU$\geq$0.75 & GIoU$\geq$0.5 & GIoU$\geq$0.3 & GIoU$\geq$0.1 & GIoU$\geq$0.0 & GIoU$\geq-\infty$ \\ \midrule
2          & 6         & 10        & 16        & 24      & 41          \\ \bottomrule
\end{tabular}
}
}
\end{table}

Additionally, we introduce Query Selection Knowledge Distillation (QSKD) framework, comprising two key components: Attention-Guided Feature Distillation (AGFD) and Local Alignment Prediction Distillation (LAPD).
For AGFD, we propose a foreground mask based on all selected queries for encoder feature distillation, ensuring a more comprehensive coverage of foreground regions.
We further investigate distillation between encoders with different numbers of layers and find that a simple transformer encoder layer as an adapter can significantly enhance AGFD performance, especially when there is no encoder layer in the student model. 
In the LAPD module, we propose an efficient and effective strategy that aligns teacher and student detector predictions, by applying GQS to both sets of predictions and then conducting bipartite matching within selected predictions. This approach ensures a robust matching process less susceptible to random noise brought by meaningless predictions, while considerably reducing the number of predictions involved in the bipartite matching process which has cubic-level
complexity for query numbers. 
As a result, LAPD can be effectively used for detectors with a large number of queries, such as the DINO detector~\cite{zhang2022dino} with $900$ queries. 

Extensive experiments demonstrate the accuracy and efficiency of our framework.
On MS-COCO dataset~\cite{coco}, our framework improves the AP of Conditional DETR~\cite{meng2021conditional} ResNet-18 from $35.8$ to $39.9$ (+$4.1$), DAB DETR~\cite{liu2022dab} ResNet-18 from $36.2$ to $41.5$ (+$5.3$) and DINO ResNet-50 from $49.0$ to $51.4$ (+$2.4$).
Our method surpasses existing approaches, achieving a new state-of-the-art in detection transformer distillation.
Furthermore, through a series of ablation studies, we demonstrate the utility of each individual module, confirming that every component plays a crucial role in achieving this advanced performance. 

In summary, our contributions can be categorized into three main aspects:
\begin{itemize}
  \item With respect to query selection for DETR distillation, we introduce the novel Group Query Selection (GQS) method, which identifies and selects the most valuable queries for distillation purposes, thereby enhancing the efficiency and effectiveness of the knowledge transfer process.
  \item We propose Knowledge Distillation via Query Selection for Detection Transformer (QSKD) framework, containing attention-guided encoder feature distillation (AGFD) and local alignment prediction distillation (LAPD).
  AGFD adeptly mitigates the challenges posed by discrepancies in encoder layer counts between student and teacher models, while LAPD facilitates the efficient establishment of local distillation pairs, optimizing the prediction alignment process.
  \item Through extensive experimentation on the COCO dataset, covering a wide range of settings, we comprehensively demonstrate the effectiveness and versatility of our proposed methods. The empirical results not only validate the superior performance of our approaches but also highlight their broad applicability across different model configurations and scenarios.
\end{itemize}

\section{Related Work}
\label{sec:relatedwork}

\subsection{Detection Transformer}
Unlike traditional detection algorithms that usually use anchors, either boxes~\cite{tang2024mi3c, ZHANG2023109664, ZHU2019106964} or points~\cite{HUA2024110541, YANG2023109627,SONG2023109278}, DETR~\cite{detr} introduces a novel approach by employing a set of trainable vectors as queries.
However, it faces challenges such as slow coverage speed and high computation costs~\cite{MO2022108899,CHEN2022108418}. In response, several subsequent works have emerged to address these limitations and improve detection accuracy.
Conditional DETR~\cite{meng2021conditional},  \dabdetr~\cite{liu2022dab}, 
AdaMixer~\cite{gao2022adamixer}, DN DETR~\cite{li2022dn}, Deformable DETR~\cite{deformable_detr}, and DINO~\cite{zhang2022dino} primarily concentrate on enhancing attention mechanisms and explicit query definition. 
Furthermore, several other studies have attempted to enhance training coverage by introducing one-to-many matching techniques during training~\cite{chen2023group, hdetr, codetr}.
Group DETR~\cite{chen2023group} introduces multiple parallel groups of queries and applies one-to-one matching within each group, allowing a single ground truth to match multiple queries in these groups.
HDETR~\cite{hdetr} proposes the inclusion of one-to-many matching in additional branches during training.
CoDETR~\cite{codetr} extends this idea further by introducing versatile label assignment methods inspired by traditional convolution-based detectors.
The characteristics of traditional detectors are becoming increasingly evident in the enhancements made to query definition, deformable attention, and the one-to-many strategy within the DETR detector.
Rather than focusing on designing new structures to enhance DETR, our approach concentrates on improving accuracy through distillation and figuring out different queries' influence on DETR distillation.

\subsection{Knowledge Distillation for Object Detection}
Knowledge distillation is a popular technique for compressing models and improving accuracy by transferring knowledge from a larger, more complex teacher model to a smaller student model. 
The concept of knowledge distillation was first introduced by \cite{kd}, which involves the student learning to mimic the soft predictions made by the teacher.
The following distillation methods can be divided into three categories: Response-Based~\cite{kd}, Feature-Based~\cite{fitnets}, and Relation-Based~\cite{rkd}.
Response-based methods leverage the insights from the logits of a large deep model.
Feature-based methods emphasize the importance of intermediate layer features~\cite{featurebase, head, Li2021c}.
Relation-based methods further incorporate relationships between features, activations, and sample pairs~\cite{de2022structural, yang2022focal}.
Besides, as the foreground area is important for the detection task, ignoring the imbalance in foreground and background will result in poor results~\cite{detection_kd}.
Various strategies for selecting informative features for distillation are discussed in the literature \cite{mimic, wang2019distillinggroundtruth, sun2020distillinggauss}. For instance, \cite{defeat} posits that features derived from regions excluding objects are also vital for training the student detector effectively. Moreover, \cite{frs} emphasizes the retrieval of important features outside bounding boxes and the removal of detrimental features within these boxes by employing FRS to identify key features.

\subsection{Knowledge Distillation for Detection Transformer}
As the Detection Transformer becomes popular, the distillation for it has emerged~\cite{huang2023teach}.
For instance, \cite{detrdistill} distills DETR at both the feature and logits levels, using similarity calculated by positive queries and feature to reweight feature distillation and employing bipartite matching to select corresponding teacher logits that guide the matched student logits. 
They also deliver the teacher's queries to the student model to get alignment predictions with teacher.
Similarly, except for logits distillation, \cite{d3etr} introduces feature distillation in the decoder, also utilizing bipartite matching to establish corresponding distillation relations.
Additionally, \cite{knowledgesampling} introduces extra queries to both the student and teacher models and distills the corresponding predictions, ensuring consistency in the distillation.
For feature distillation in the encoder, these works concentrate on positive queries, ignoring other queries not matched with objects by bipartite matching but identifying objects.
For logits or feature distillation in the decoder, these works either conduct bipartite matching between all teacher and student queries or introduce new queries to both the teacher and student model to establish consistency, not distinguishing the role of different queries and introducing some unnecessary computation cost.

\section{Method}
\label{sec:method}
\begin{figure*}[t]
  \centering
  \includegraphics[width=0.99\linewidth, ]
  {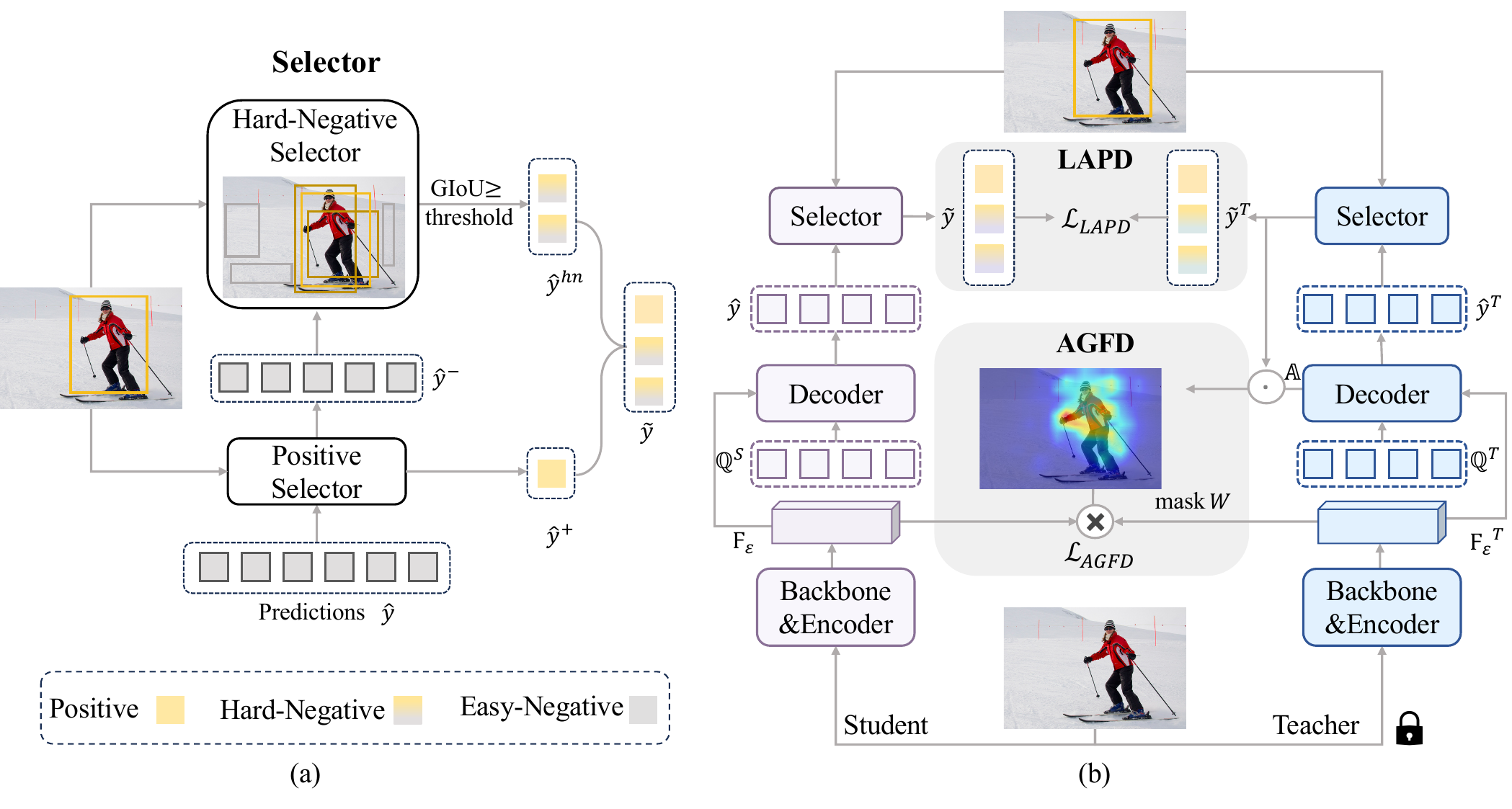}
  \caption{
    The framework of our Query Selection Knowledge Distillation (QSKD):
    (a) Group Query Selector: Initially, positive queries are identified through bipartite matching with ground truth boxes. Subsequently, we rank all negative queries based on their GIoU with ground truth boxes and select the hard-negative queries whose GIoU is bigger than the threshold. These are then combined with the positive queries to finalize our selection.
    (b) Our distillation architecture utilizing the Group Query Selector (GQS) includes Attention-Guided Feature Distillation (AGFD) and Local Alignment Prediction Distillation (LAPD). Both the teacher and student models are Detection Transformers. To simplify our explanation, details of the original training phase supervision are omitted.
  }
  \label{fig:framework}
\end{figure*}

The section begins with a brief overview of transformer-based detectors, detailed in \Cref{subsec:preliminaries}. 
Subsequent to this foundational overview, \Cref{subsec:queryselection} delves into a comprehensive analysis of our Group-Query Selection (GQS) method. 
We then progress to the Query Selection Knowledge Distillation (QSKD) framework, which is comprised of two essential components: Attention-Guided Feature Distillation (AGFD) and Local Aligned Prediction Distillation (LAPD). 
These modules are thoroughly expounded in \Cref{subsec:encdistill} and \Cref{subsec:decdistill}, respectively.
Finally, the computation and implications of the total loss, which integrates contributions from both AGFD and LAPD alongside other loss components, are comprehensively described in \Cref{subsec:total_loss}.

\subsection{Preliminaries}
\label{subsec:preliminaries}

A typical transformer-based detector, such as DETR~\cite{detr}, consists of three principal components: a backbone $\mathcal{B}$, a transformer encoder $\mathcal{E}$, and a transformer decoder $\mathcal{D}$. 
Given an input image $\mathbf{I} \in \mathbb{R}^{H \times W \times 3}$, the backbone $\mathcal{B}$ functions to extract the image features $\mathbf{F}_\mathcal{B}$.
These features are then refined via the encoder $\mathcal{E}$, which employs self-attention mechanisms, leading to $\mathbf{F}_\mathcal{E} = \mathcal{E}(\mathbf{F}_\mathcal{B})$.

The decoder $\mathcal{D}$ processes $\mathbf{F}_\mathcal{E}$ alongside a set of learnable object queries $\mathbf{Q} = \{q_i\}_{i=1}^{N_q}\subset \mathbb{R}^{d}$ as inputs, where $N_q$ is the number of queries and $d$ is the embedding dimension. 
For each query $q_i$, the decoder $\mathcal{D}$ generates predictions $\hat{y}_i=(\hat{c}_i, \hat{b}_i)$, with $\hat{c}_i$ and $\hat{b}i$ representing the classification and bounding box predictions, respectively.
The final output of $\mathcal{D}$ is represented as $\hat{y} = \mathcal{D}(\mathbf{Q}, \mathbf{F}_\mathcal{E}) = \{\hat{y}_i\}^{N_q}_{i=1}$.

The ground truth objects are denoted as $y^\text{gt} = \{y_i\}^{N_\text{gt}}_{i=1}$. 
To align with the number of queries $N_q$, $y^\text{gt}$ is padded with $\varnothing$, which indicates the absence of an object, resulting in the targets $y = {y_i}^{N_q}{i=1}$.
Bipartite matching is performed between $\hat{y}$ and $y$, searching for an optimal permutation:
\begin{equation}
  \hat{\sigma} = \arg\min_{\sigma} \sum_{i=1}^{N_q} \mathcal{L}_\text{match}(y_{i}, \hat y_{\sigma(i)}),
  \label{eq:binmatch}
\end{equation} 
where $\sigma$ represents any permutation of size $N_q$, $\sigma(i)$ denotes the $i$-th element in $\sigma$, and $\mathcal{L}_\text{match}$ refers to the matching cost which often contains regression cost and classification cost.

The overall loss is defined as follows:
\begin{equation}
    \mathcal{L}_\text{gt}(y, \hat{y}) = \sum_{i=1}^{N_q} \left[ 
        \lambda_{\text{cls}}\mathcal{L}_{\text{cls}}(c_i, \hat{c}_{\hat{\sigma}(i)})
        + \mathbbm{1}\{{c_i\neq\varnothing}\} \cdot \lambda_{\text{box}}\mathcal{L}_{\text{box}}(b_i, \hat{b}_{\hat{\sigma}(i)})
     \right],
    \label{eq:gtloss}
\end{equation}
where $\mathcal{L}_{\text{cls}}$ represents the classification loss and $\mathcal{L}_{\text{box}}$ denotes the regression loss.
$\lambda_{\text{cls}}$ and $\lambda_{\text{box}}$ are the respective loss weights.
$\mathbbm{1}$ is an indicator function.

\subsection{Group Query Selection}
\label{subsec:queryselection}

The key to query selection lies in identifying the most informative queries for distillation.
A basic approach to query selection is using the bipartite matching algorithm.
This algorithm, when applied between the predictions $\hat{y}$ and targets $y$, enables the division of $\hat{y}$ into two groups: positive $\hat{y}^+ = \{\hat{y}_{\hat{\sigma}(i)} | c_i \ne \varnothing\}$ and negative $\hat{y}^- = \{\hat{y}_{\hat{\sigma}(i)} | c_i = \varnothing\}$.
The indices for these groups are denoted as $\mathbb{I}^+ = \{\hat{\sigma}(i) | c_i \ne \varnothing\}$ and $\mathbb{I}^- = \{\hat{\sigma}(i) | c_i = \varnothing\}$, respectively.
Owing to direct correspondence between queries and predictions, the queries $\mathbf{Q}$ can be similarly segmented into positive queries $\mathbf{Q}^+$ and negative queries $\mathbf{Q}^-$, matching with $y_\text{gt}$ and $\varnothing$, respectively.
While it may seem intuitive to use only positive queries for distillation and discard the negative ones~\cite{detrdistill}, we have observed that many negative queries are also informative, as depicted in \Cref{fig:queries}.
Therefore, we propose a Group Query Selection (GQS) method to retrieve as many informative queries as possible, if not all.

As illustrated in \Cref{fig:framework}~(a), our proposed method comprises a positive selector and a hard-negative selector.
The positive selector straightforwardly conducts bipartite matching between the set of predictions \(\hat{y}\) and the ground truth objects \(y^\text{gt}\).
For each negative prediction $\hat{y}_i^-$, the hard-negative selector begins by calculating its Generalized Intersection over Union (GIoU) relative to the ground truth objects $y^\text{gt}$.
Subsequently, $\hat{y}_i^-$ is associated with the ground truth object exhibiting the highest GIoU, defining this highest GIoU as the GIoU metric $G_i$ for $\hat{y}_i^-$.
This process creates $N_\text{gt}$ clusters of negative predictions $\hat{y}^-$.
Within each cluster, the predictions with the largest GIoU higher than the threshold are classified as hard-negative predictions, while the rest are considered easy-negative. 
The assumption is that the queries associated with hard-negative predictions, termed hard-negative queries, are also rich in information valuable for the distillation process.
Hence, both positive and hard-negative queries for distillation. 
For clarity, the indices of the chosen hard-negative queries are denoted as $\mathbb{I}^\text{hn}$, the corresponding queries as $\mathbf{Q}^\text{hn} = \{q_i | i \in \mathbb{I}^\text{hn}\}$, and their GIoU metrics as $\mathbf{G}^\text{hn} = \{G_i | i \in \mathbb{I}^\text{hn}\}$.

The final output of GQS is defined as:
\begin{equation}
\begin{split}
\tilde{\mathbb{I}} &= \mathbb{I}^+ \cup \mathbb{I}^\text{hn}, \\
\tilde{y} &= \hat y^+ \cup \hat{y}^\text{hn}.
\end{split}
\end{equation}
 
\subsection{Attention-Guided Feature Distillation}
\label{subsec:encdistill}

Previous research in knowledge distillation for object detection has shown that foreground regions are more informative than background areas~\cite{detection_kd}.
For instance, DETRDISTILL emphasizes the foreground using a similarity matrix between positive queries $\mathbf{Q}^+$ and image features.
However, we posit that an exclusive focus on $\mathbf{Q}^+$ might overlook substantial parts of the foreground.
In light of this, our Group Query Selection (GQS) algorithm puts forward \(\tilde{\mathbf{Q}}\) as a more apt choice, which includes both positive and hard-negative queries, offering a broader scope for foreground emphasis.

In line with this approach, we introduce Attention-Guided Feature Distillation (AGFD).
As shown in \Cref{fig:framework}~(b), this technique leverages the attention matrix between $\tilde{\mathbf{Q}}$ and the image features $\mathbf{F}_\mathcal{E}$ to generate a foreground mask.
We define $\mathbb{A} = \{\mathbf{A}_i\}_{i=1}^{N_q}$ as the attention matrix in the first layer of the teacher decoder $\mathcal{D}^T$, with
$\mathbf{A}_i$ representing the attention matrix between the $i$-th query $q_i$ and $\mathbf{F}_\mathcal{E}$.
The foreground mask is formulated as:
\begin{equation}
    \mathbf{W} = \frac{1}{|\tilde{\mathbb{I}}|} \sum_{i \in \tilde{\mathbb{I}}} (1 + G_i)\cdot \mathbf{A}_i,
\end{equation}
where $\tilde{\mathbb{I}}$ denotes the indices of the selected queries, $|\mathbb{I}|$ the number of $\tilde{\mathbb{I}}$,  and $G_i$ the GIoU metric of the $i$-th query.
It's important to note that $\mathbf{W}$ is a weighted average of $\mathbb{A}$, with the weights being the GIoU metrics of the selected queries.

The AGFD loss of our framework is formulated as:
\begin{equation}
  \mathcal{L}_\text{AGFD} = D(
    \beta(\mathbf{W} \cdot \mathbf{F}_\mathcal{E}^T), 
    \beta(\mathbf{W} \cdot \mathbf{F}_\mathcal{E})
  ),
  \label{eq:encloss}
\end{equation}
where $D(\cdot)$ represents the MSE loss, $\beta(\cdot)$ is a batch normalization layer, $\mathbf{F}_\mathcal{E}^T$ and $\mathbf{F}_\mathcal{E}$ are the encoder outputs of the teacher and student detectors, respectively.

\begin{figure*}[t]
\centering
  \begin{subfigure}{0.23\textwidth}
    \includegraphics[width=1\textwidth]{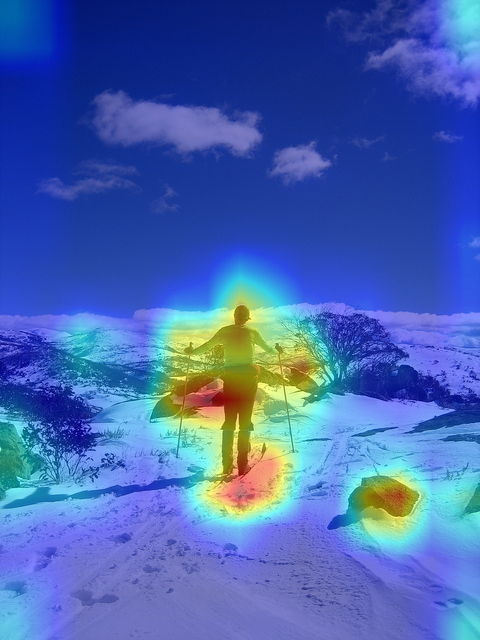}
    \caption{No Encoder}
    \label{fig:fpn}
  \end{subfigure}
  \begin{subfigure}{0.23\textwidth}
    \includegraphics[width=1\textwidth]{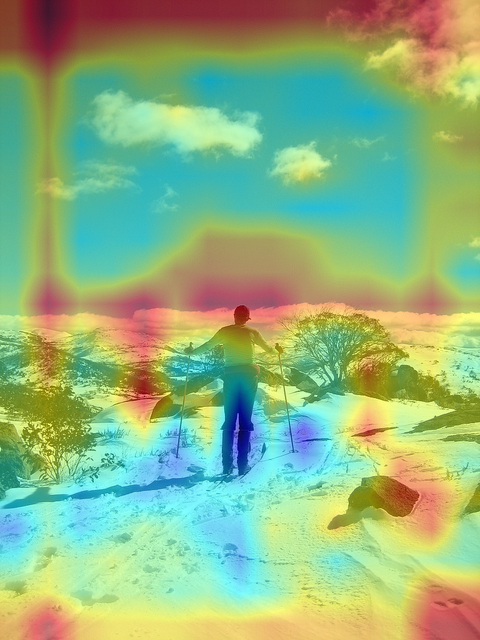}
    \caption{Single Layer}
    \label{fig:ada}
  \end{subfigure}
  \begin{subfigure}{0.23\textwidth}
    \includegraphics[width=1\textwidth]{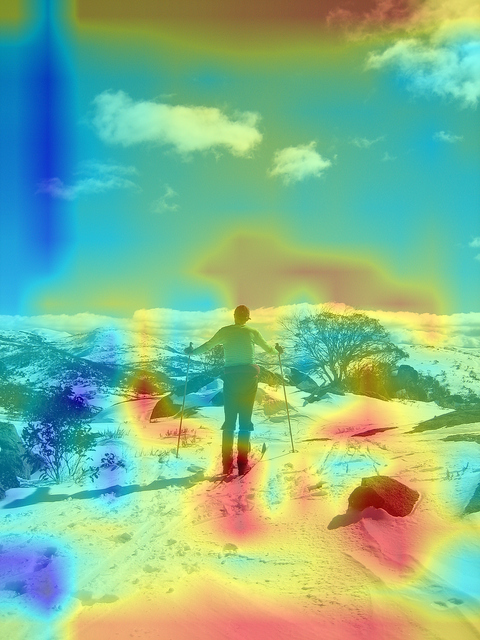}
    \caption{Six Layers}
    \label{fig:enc}    
  \end{subfigure}
  \caption{
    We visualize image features from detectors with no encoder, $1$ encoder layer, and $6$ encoder layers. 
    It highlights a transition in activation patterns when adding a single encoder layer, while the difference between $1$ and $6$ layers is less pronounced.
  }
  \label{fig:adapter}
\end{figure*}

We further explore encoder feature distillation across encoders with varying numbers of layers and reveal an interesting finding: utilizing a single encoder layer as an adapter significantly enhances distillation performance.
Notably, this approach remains effective even when the number of encoder layers is reduced to zero.
This empirical observation is supported by the visualizations in \Cref{fig:adapter}, which show that the output feature of a six-layer encoder markedly differs from that of a CNN-based backbone (\ie, no encoder).
Consequently, directly applying the foreground mask \(\mathbb{W}\) to the student's image feature may not effectively highlight the regions requiring distillation.
However, by integrating an encoder layer as an adapter, the student's image feature becomes more akin to that of the teacher.
This alignment allows the foreground mask $\mathbb{W}$ to more accurately highlight the regions valuable for distillation, thereby enhancing the effectiveness of our AGFD.
Then the distillation loss will change to :

\begin{equation}
\mathcal{L}_\text{AGFD} = D(
    \beta(\mathbf{W} \cdot \mathbf{F}_\mathcal{E}^T), 
    \beta(\mathbf{W} \cdot  \mathcal{E}_{a}(\mathbf{F}_{a}), )
  ),
\end{equation}
where $\mathcal{E}a(\cdot)$ represents a transformer encoder layer serving as an adaptor, while $\mathbf{F}{a}$ denotes the feature forwarded to the student decoder. This feature, $\mathbf{F}{a}$, could either be $\mathbf{F}_\mathcal{B}$ or a feature output from various encoder layers.

\subsection{Local Aligned Prediction Distillation}
\label{subsec:decdistill}

The goal of our Local Aligned Prediction Distillation (LAPD) is to transfer knowledge from the teacher's predictions $\hat{y}^T$ to the student's predictions $\hat{y}$. 
A critical phase in this process is the establishment of a correspondence between $\hat{y}^T$ and $\hat{y}$. 
While bipartite matching has been the predominant method used in previous approaches for this purpose, it presents considerable challenges, both in terms of both precision and efficiency.

The precision issue in bipartite matching stems from the inherent diversity of negative predictions.
Owing to this diversity, bipartite matching may erroneously pair unrelated predictions, thereby introducing noise into the distillation process. 
Regarding computational efficiency, the cost of bipartite matching escalates quadratically with the number of queries $N_q$. 
This poses a significant problem for detectors with a large number of queries. 
For instance, in the case of the DINO detector, which operates with $900$ queries, the computational load of executing bipartite matching is $5$ times more than the baseline approach. 
Such a marked increase in computational demands renders bipartite matching unfeasible in such scenarios.

Our proposed Local Aligned Prediction Distillation (LAPD) method effectively tackles both the precision and efficiency issues through query sampling.
Initially, the Group Query Selection (GQS) is applied to both the teacher's predictions $\hat{y}^T$ and the student's predictions $\hat{y}$.
For each ground truth object $y_i \in y^\text{gt}$, we match the corresponding positive predictions, $\hat{y}^T_{\hat{\sigma}(i)}$ and $\hat{y}_{\hat{\sigma}(i)}$.
Following this, bipartite matching is conducted within the selected negative predictions associated with $y_i$.
Notably, this matching is restricted to the selected queries, which aids in effectively suppressing noise.

This approach results in a partial match between $\hat{y}^T$ and $\hat{y}$, significantly reducing computation cost.
The partial match is represented as $\{(\hat{y}^T_{\hat{\sigma}^T_\text{LAPD}(i)}, \hat{y}_{\hat{\sigma}_\text{LAPD}(i)})\}_{i=1}^{N_\text{LAPD}}$, where $\hat{\sigma}^T_\text{LAPD}$ and $\hat{\sigma}_\text{LAPD}$ are two permutatins of size $N_q$, and $N_\text{LAPD}$ is the number of matched query pairs.
Distillation is then performed between these matched predictions as follows:
\begin{equation}
    \mathcal{L}_{\text{LAPD}} = \sum_{i=1}^{N_\text{LAPD}}
    \left[
        \lambda_{\text{cls}} \mathcal{L}_{\text{cls}}(\hat{c}^T_{\hat{\sigma}^T_\text{LAPD}(i)}, \hat{c}_{\hat{\sigma}_\text{LAPD}(i)}) 
        + \lambda_{\text{box}} \mathcal{L}_{\text{box}}(\hat{b}^T_{\hat{\sigma}^T_\text{LAPD}(i)}, \hat{b}_{\hat{\sigma}_\text{LAPD}(i)}) 
    \right],
\label{eq:decloss}
\end{equation}
where $\mathcal{L}_{\text{cls}}$ represents the classification loss and $\mathcal{L}_{\text{box}}$ denotes the regression loss.
$\lambda_{\text{cls}}$ and $\lambda_{\text{box}}$ are the respective loss weights.

\subsection{Total Loss}
\label{subsec:total_loss}

The overall training loss for our Query Selection Knowledge Distillation framework is formulated as follows:
\begin{equation}
  \mathcal{L} = \mathcal{L}_{\text{gt}} + \lambda_\text{AGFD}\mathcal{L}_\text{AGFD} + \lambda_\text{LAPD}\mathcal{L}_\text{LAPD},
\end{equation}
where $\mathcal{L}_{\text{gt}}$ represents the original ground-truth loss (\Cref{eq:gtloss}), $\mathcal{L}_\text{AGFD}$ is the AGFD loss (\Cref{eq:encloss}), and $\mathcal{L}_\text{LAPD}$ denotes the LAPD loss (\Cref{eq:decloss}). The parameters $\lambda_\text{AGFD}$ and $\lambda_\text{LAPD}$ are the respective weights assigned to the AGFD and LAPD loss components. 

\section{Experiments}
\label{sec:experiments}

\subsection{Experiment Setup}
\paragraph{Dataset} We conduct experiments on the MS-COCO 2017 dataset~\cite{coco}, which contains $118,287$ images for training and $5,000$ images for validation.
We report mean Average Precision (AP) as an evaluation metric and AP under different scales, \ie, AP$\rm _S$, AP$\rm _M$, AP$\rm _L$.

\paragraph{Implementation Details}We conducted our experiments on $32$GB NVIDIA Tesla V100 GPUs, with a total batch size of $16$. 
Our methodology is grounded in the official versions of well-regarded models: Conditional DETR~\cite{meng2021conditional}, DAB DETR~\cite{liu2022dab}, and DINO~\cite{zhang2022dino}, recognized for their popularity and rapid convergence. 
Emphasizing the modular nature of our distillation approach, we adhered strictly to the original configuration of each model, including its hyperparameters, learning rate schedule, and optimizer settings. 
The training of the teacher and student models was conducted for $50/36/12$ epochs using the AdamW optimizer, characterized by a weight decay of $0.0001$.
We set the loss weights, $\lambda_\text{AGFD}$ for the encoder and $\lambda_\text{LAPD}$ for the decoder, at $50$ and $1$, respectively, to ensure balanced training dynamics.
To quantitatively assess the efficiency of our method, we measured the Frames Per Second (FPS) during inference on a $32$ GB NVIDIA Tesla V$100$ GPU, providing a practical benchmark of our model's performance in real-world conditions.

\subsection{Main Results}
\begin{table*}[htbp]
\centering
\caption{Experimental results of our QSKD on different detectors and backbones. For each detector and backbone, the first row is the baseline, and the last row corresponds to our results. We also include some other DETR distillation method results for comparison.}
\resizebox{1.0\textwidth}{!}{%
\setlength{\extrarowheight}{3pt}{
\setlength{\tabcolsep}{2.mm}{
\begin{tabular}{@{}cccccccccc@{}}
\toprule
Teacher                                                                                            & Method                          & Queries              & Epoch               & AP            & AP$\rm_{50}$  & AP$\rm_{75}$  & AP$\rm _S$    & AP$\rm _M$    & AP$\rm _L$    \\ \midrule
\multirow{9}{*}{\begin{tabular}[c]{@{}c@{}}Conditional DETR~\cite{meng2021conditional}\\ ResNet-101\\ 42.8(50e)\end{tabular}} & ResNet-18 (S)                        & 300                  & \multirow{4}{*}{50} & 35.8          & 56.2          & 37.5          & 16.1          & 38.5          & 54.5          \\
                                                                                                   & D3ETR~\cite{d3etr}               & 300+300              &                     & 39.6          & -             & -             & 18.8          & 42.6          & 59.2          \\
                                                                                                   & \multirow{2}{*}{Ours}            & \multirow{2}{*}{300} &                     & \cellcolor{gray!25}\textbf{39.9} & \cellcolor{gray!25}\textbf{60.3} & \cellcolor{gray!25}\textbf{42.1} & \cellcolor{gray!25}\textbf{19.3} & \cellcolor{gray!25}\textbf{43.0} & \cellcolor{gray!25}\textbf{59.2} \\
                                                                                                   &                                  &                      &                     &\cellcolor{gray!25} \cellcolor{gray!25}\textbf{+4.1} & \cellcolor{gray!25}\textbf{+4.1} & \cellcolor{gray!25}\textbf{+4.6} & \cellcolor{gray!25}\textbf{+3.2} & \cellcolor{gray!25}\textbf{+4.5} & \cellcolor{gray!25}\textbf{+4.7} \\ \cmidrule(l){2-10} 
                                                                                                   & ResNet-50 (S)                        & 300                  & \multirow{5}{*}{50} & 40.9          & 61.7          & 43.3          & 20.6          & 44.3          & 59.3          \\
                                                                                                   & D3ETR~\cite{d3etr}               & 300+300              &                     & 43.3          & -             & -             & 22.3          & 46.9          & 62.1          \\
                                                                                                   & DETRDISTILL~\cite{detrdistill}   & 300+300              &                     & 42.9          & -             & -             & 21.6          & 46.5          & \textbf{62.2} \\
                                                                                                   & \multirow{2}{*}{Ours}           & \multirow{2}{*}{300} &                    &\cellcolor{gray!25} \textbf{43.7} & \cellcolor{gray!25}\textbf{64.4} & \cellcolor{gray!25}\textbf{46.8} & \cellcolor{gray!25}\textbf{23.6} & \cellcolor{gray!25}\textbf{47.3} & \cellcolor{gray!25}61.9          \\
                                                                                                   &                                  &                    &                     &\cellcolor{gray!25} \textbf{+2.8} & \cellcolor{gray!25}\textbf{+2.7} & \cellcolor{gray!25}\textbf{+3.5} & \cellcolor{gray!25}\textbf{+3.0} & \cellcolor{gray!25}\textbf{+3.0} & \cellcolor{gray!25}+2.6          \\ \midrule
\multirow{3}{*}{\begin{tabular}[c]{@{}c@{}}DAB DETR~\cite{liu2022dab}\\ ResNet-101\\ 43.5(50e)\end{tabular}}         & ResNet-18 (S)                        & \multirow{3}{*}{300} & \multirow{3}{*}{50} & 36.2          & 56.1          & 37.9          & 16.9          & 39.0          & 53.5          \\
                                                                                                   & \multirow{2}{*}{Ours}            &                      &                     & \cellcolor{gray!25}\textbf{41.5} & \cellcolor{gray!25}\textbf{61.2} & \cellcolor{gray!25}\textbf{44.2} & \cellcolor{gray!25}\textbf{20.9} & \cellcolor{gray!25}\textbf{44.8} & \cellcolor{gray!25}\textbf{60.0} \\
                                                                                                   &                                  &                      &                     &\cellcolor{gray!25}\textbf{+5.3} & \cellcolor{gray!25}\textbf{+5.1} & \cellcolor{gray!25}\textbf{+6.3} & \cellcolor{gray!25}\textbf{+4.0} & \cellcolor{gray!25}\textbf{+5.8} & \cellcolor{gray!25}\textbf{+6.5} \\ \midrule
\multirow{4}{*}{\begin{tabular}[c]{@{}c@{}}DAB DETR~\cite{liu2022dab}\\ ResNet-50\\ 42.1(50e)\end{tabular}}          & ResNet-18 (S)                        & 300                  & \multirow{4}{*}{50} & 36.2          & 56.1          & 37.9          & 16.9          & 39.0          & 53.5          \\
                                                                                                   & KD-DETR~\cite{knowledgesampling} & 300+600              &                     & 41.4          & 61.4          & 44.2          & \textbf{21.2} & 44.7          & 58.7          \\
                                                                                                   & \multirow{2}{*}{Ours}            & \multirow{2}{*}{300} &                     &\cellcolor{gray!25}\textbf{41.5} & \cellcolor{gray!25}\textbf{61.4} & \cellcolor{gray!25}\textbf{44.2} & \cellcolor{gray!25}20.7          & \cellcolor{gray!25}\textbf{44.8} & \cellcolor{gray!25}\textbf{61.0} \\
                                                                                                   &                                  &                      &                     & \cellcolor{gray!25}\textbf{+5.3} & \cellcolor{gray!25}\textbf{+5.3} & \cellcolor{gray!25}\textbf{+6.3} & \cellcolor{gray!25}+3.8          & \cellcolor{gray!25}\textbf{+5.8} & \cellcolor{gray!25}\textbf{+7.5} \\ \midrule
\multirow{4}{*}{\begin{tabular}[c]{@{}c@{}}DINO~\cite{zhang2022dino}\\ ResNet-50\\ 50.9(36e)\end{tabular}}              & ResNet-18 (S)                        & 900                  & \multirow{4}{*}{12} & 44.0          & 61.2          & 48.1          & 27.4          & 46.9          & 56.9          \\
                                                                                                   & KD-DETR~\cite{knowledgesampling} & 900+1800             &                     & 45.4          & 62.2          & 49.3          & 27.3          & 48.2          & 49.0          \\
                                                                                                   & \multirow{2}{*}{Ours}            & \multirow{2}{*}{900} &                     & \cellcolor{gray!25}\textbf{47.3} & \cellcolor{gray!25}\textbf{63.9} & \cellcolor{gray!25}\textbf{51.8} & \cellcolor{gray!25}\textbf{29.9} & \cellcolor{gray!25}\textbf{50.3} & \cellcolor{gray!25}\textbf{61.5} \\
                                                                                                   &                                  &                      &                     & \cellcolor{gray!25}\textbf{+3.3} & \cellcolor{gray!25}\textbf{+2.7} & \cellcolor{gray!25}\textbf{+3.7} & \cellcolor{gray!25}\textbf{+2.5} & \cellcolor{gray!25}\textbf{+3.4} & \cellcolor{gray!25}\textbf{+4.6} \\ \midrule
\multirow{3}{*}{\begin{tabular}[c]{@{}c@{}}DINO~\cite{zhang2022dino}\\ Swin-L\\ 58.0(36e)\end{tabular}}                 & ResNet-50 (S)                        & \multirow{3}{*}{900} & \multirow{3}{*}{12} & 49.0          & 66.6          & 53.5          & 32.0          & 52.3          & 63.0          \\
                                                                                                   & \multirow{2}{*}{Ours}            &                      &                     & \cellcolor{gray!25}\textbf{51.4} & \cellcolor{gray!25}\textbf{68.5} & \cellcolor{gray!25}\textbf{55.9} & \cellcolor{gray!25}\textbf{33.7} & \cellcolor{gray!25}\textbf{54.9} & \cellcolor{gray!25}\textbf{67.4} \\
                                                                                                   &                                  &                      &                     & \cellcolor{gray!25}\textbf{+2.4} & \cellcolor{gray!25}\textbf{+1.9} & \cellcolor{gray!25}\textbf{+2.4} & \cellcolor{gray!25}\textbf{+1.7} & \cellcolor{gray!25}\textbf{+2.6} & \cellcolor{gray!25}\textbf{+4.4} \\ \bottomrule
\end{tabular}
}
}
}
\label{tab:dismain}
\end{table*}
In this investigation, we evaluate the performance of our distillation methodology across three Detection Transformer variants: Conditional DETR and DAB DETR, both single-scale, one-stage models, and DINO, a multi-scale, two-stage model noted for its state-of-the-art performance. 
We employed a range of teacher backbones, including ResNet-101~\cite{resnet}, ResNet-50, and Swin-Large~\cite{swin}, to guide student models built on ResNet-18 and ResNet-50 backbones. The training was configured to last $50$ epochs for the Conditional DETR and DAB DETR student models, and $12$ epochs for the DINO student models.

Our experiments were structured to explore distillation effectiveness under varied conditions, including identical teacher models guiding different student models and the use of diverse teacher models for the same student backbone.
Initial experiments utilizing a Conditional DETR ResNet-101 teacher model demonstrated significant performance boosts, with ResNet-18 and ResNet-50 student backbones achieving improvements of $4.1$ AP and $2.8$ AP, respectively. Notably, the Conditional DETR model with a ResNet-50 student even exceeded its teacher's performance by $0.9$ AP, as detailed in Table~\ref{tab:dismain}.
Further experimentation with DAB DETR teachers (ResNet-101 and ResNet-50) training a ResNet-18 student yielded a uniform enhancement of $5.3$ AP across configurations. This uniformity in improvement underscores the versatility of our distillation approach.
Additionally, our methodology significantly bolstered the performance of the DINO detector, manifesting in a $3.3$ AP increase when utilizing a ResNet-50 teacher with a ResNet-18 student.
An attempt to integrate a Swin-L backbone for the teacher model further validated our method's effectiveness, contributing to a $2.4$ AP gain for a ResNet-50 based DINO student model.
These findings illustrate the broad applicability and robustness of our distillation method across various Detection Transformer configurations and training scenarios. By demonstrating substantial improvements in model performance without the necessity for additional queries or complex modifications, our approach marks a significant advancement in the efficiency and effectiveness of model training in the object detection domain.

In the landscape of distillation strategies for object detection models, our proposed method showcases exemplary performance, surpassing established benchmarks. Specifically, when juxtaposed with DETRDISTILL~\cite{detrdistill} within a Conditional ResNet-50 framework, our approach yields a notable improvement, achieving a $43.7$ AP compared to their reported $42.9$ AP. Remarkably, this enhancement is realized \textbf{without incorporating additional teacher queries} into the student decoder during the training phase.
Moreover, our method demonstrates its robustness by outperforming D3ETR~\cite{d3etr} across both Conditional ResNet-18 and ResNet-50 configurations, with respective AP gains of $0.3$ and $0.4$. This achievement is underscored by a training process characterized by both increased speed and simplicity. Against KD-DETR~\cite{knowledgesampling}, our strategy secures a substantial lead, surpassing the DINO ResNet-18 student model, guided by a DINO ResNet-50 teacher, by an impressive margin of $1.9$ AP.
A critical distinction of our method lies in its strategic utilization of the inherent knowledge embedded within the original student queries, eschewing the need for the integration of extra queries for distillation purposes. This approach not only simplifies the distillation process but also enhances its efficiency, offering a streamlined pathway to knowledge transfer. It is pivotal to acknowledge that our method does not preclude the potential benefits of distillation involving additional queries, suggesting a complementary rather than conflicting relationship with such strategies.
This comparative analysis emphasizes the superiority of our distillation method, highlighting its potential to significantly advance the efficiency and effectiveness of object detection model training. Through a meticulous blend of innovation and strategic insight, our method sets a new benchmark in the domain, promising to catalyze further advancements in model distillation techniques.
\begin{table}[]
\centering
\caption{Distillation performance of QSKD on Detection Transformers with identical teacher and student backbones}
\setlength{\extrarowheight}{2pt}{
\begin{tabular}{@{}cccccc@{}}
\toprule
Detector                                                                                   & Setting          & AP            & AP$\rm _S$    & AP$\rm _M$    & AP$\rm _L$    \\ \midrule
\multirow{2}{*}{\begin{tabular}[c]{@{}c@{}}Conditional DETR~\cite{meng2021conditional}\\ ResNet-50(50e)\end{tabular}} & baseline/student & 40.9          & 20.6          & 44.3          & 59.3          \\
                                                                                           & Ours             & \textbf{42.2} & \textbf{21.5} & \textbf{45.4} & \textbf{60.7} \\ \midrule
\multirow{2}{*}{\begin{tabular}[c]{@{}c@{}}DAB DETR~\cite{liu2022dab}\\ ResNet-50(50e)\end{tabular}}         & teacher/student  & 42.2          & 21.5          & 45.7          & 60.3          \\
                                                                                           & Ours             & \textbf{42.9} & \textbf{22.4} & \textbf{46.4} & \textbf{61.0} \\ \midrule
\multirow{2}{*}{\begin{tabular}[c]{@{}c@{}}DINO~\cite{zhang2022dino}\\ ResNet-50(12e)\end{tabular}}             & teacher/student  & 49.0          & 32.0          & 52.3          & 63.0          \\
                                                                                           & Ours             & \textbf{49.7} & \textbf{32.7} & \textbf{53.1} & \textbf{63.8} \\ \bottomrule
\end{tabular}%
}

\label{tab:selfdis}
\end{table}
\subsection{Self-Distillation Results}
Our method proves effective even when the teacher and student backbones are identical. In particular, we adopted ResNet-50 as the common backbone for both teacher and student models in our experiments. The results, as presented in Table~\ref{tab:selfdis}, affirm the efficacy of our approach: our model achieves an enhancement of $1.3$ AP for Conditional DETR, $0.7$ AP for DAB DETR, and $0.7$ AP for DINO. 
This demonstrates the versatility of our method, capable of delivering substantial improvements irrespective of whether the teacher and student models share the same backbone.

\subsection{Ablation Studies}
\begin{table}[]
\begin{center}
\captionof{table}{Effectiveness of each component in QSKD. The gray line indicates we use the teacher's parameter to initialize the student}
\label{tab:abl}
\setlength{\extrarowheight}{2pt}{
\begin{tabular}{cccccccc}
\toprule
AGFD & LAPD & AP            & \multicolumn{1}{l}{AP$\rm_{50}$} & \multicolumn{1}{l}{AP$\rm_{75}$} & AP$\rm_S$     & AP$\rm_M$     & AP$\rm _L$    \\ \midrule
\multicolumn{1}{l}{}       & \multicolumn{1}{l}{}       & 35.8          & 56.2                             & 37.5                             & 16.1          & 38.5          & 54.5          \\
\checkmark                 & \multicolumn{1}{l}{}       & 39.4          & 60.0                             & 41.7                             & 18.3          & 42.7          & 58.2          \\
\multicolumn{1}{l}{}       & \checkmark                 & 37.8          & 58.2                             & 39.7                             & 17.8          & 40.7          & 56.7          \\
\checkmark                 & \checkmark                 & 39.6          & 60.0                             & 41.9                             & 18.9          & 42.8          & 58.5          \\
\rowcolor[HTML]{EFEFEF} 
\checkmark                 & \checkmark                 & \textbf{39.9} & \textbf{60.3}                    & \textbf{42.1}                    & \textbf{19.3} & \textbf{43.0} & \textbf{59.2} \\ \bottomrule
\end{tabular}
}
\end{center}

\end{table}
In this section, we conduct ablation experiments on each component of QSKD: attention-guided feature distillation, local alignment prediction distillation, and encoder-free distillation.
We specifically employ Conditional DETR ResNet-101 as the teacher model and Conditional DETR ResNet-18 as the student model.
We train the student model for $50$ epochs in the main ablation experiments. 
In the more focused component ablation studies, the student model underwent training for $12$ epochs, with a scheduled reduction in the learning rate occurring in the $11$th epoch.
\subsubsection{Main Ablation} 
To study the impact of each component in our method, we report the performance of each module in Table~\ref{tab:abl}. Our experimental results reveal that applying AGFD alone results in a performance gain of $3.6$ AP while applying LAPD alone leads to a gain of $2.0$ AP.
Since the encoder and decoder components remain the same, we preload the parameters of the teacher directly into the student, which can be viewed as a form of feature distillation. It not only accelerates the convergence of the model~\cite{icd} but also improves accuracy by $0.3$ AP, as shown in the last row in Table~\ref{tab:abl}. 
Combining all these elements gives us a solid $4.1$ AP improvement over the Conditional DETR ResNet-18 baseline, confirming the effectiveness of our approach.

\begin{table}[]
\centering
\caption{Ablation study for Attention-Guided Feature Distillation with different $k$ settings and GIoU reweighting.}
\setlength{\extrarowheight}{2pt}{
\setlength{\tabcolsep}{4mm}{
\begin{tabular}{@{}ccccc@{}}
\toprule
Setting                            & AP            & AP$\rm _S$    & AP$\rm _M$    & AP$\rm _L$    \\ \midrule
Baseline                           & 32.4          & 12.6          & 34.2          & 50.3          \\
$\mathbf{Q}^+$                     & 32.7          & 13.7          & 34.9          & 50.2          \\
$\mathbf{Q}^{en}$                                            & 31.6          & 13.2          & 34.0          & 47.6          \\
$\tilde{\mathbf{Q}}_{thr=0.5}$         & 33.4          & 13.9          & 35.7          & 51.5          \\
$\tilde{\mathbf{Q}}_{thr=0.1}$        & 33.6          & 13.3          & 35.9          & 51.3          \\
$\tilde{\mathbf{Q}}_{thr=0.0}$        & 33.7          & 13.4          & 35.9          & \textbf{52.2}          \\
$\tilde{\mathbf{Q}}_{thr=0.1}$ + GIoU & 33.7          & 13.4          & 36.1          & 52.0          \\
$\tilde{\mathbf{Q}}_{thr=0.0}$ + GIoU & \textbf{33.9} & \textbf{13.7} & \textbf{36.5} & 51.7 \\ 
\bottomrule
\end{tabular}
}
}
\label{tab:abl_enc}
\end{table}
\subsubsection{Ablations on the AGFD} 
In this research, we explore the effective application of cross-attention to enhance encoder feature distillation within DETR (Detection Transformer) models. 
Our methodology encompasses a variety of encoder feature distillation strategies, incorporating different Generalized Intersection Over Union (GIoU) thresholds for query selection and GIoU-based rescoring approaches.
As a foundation for comparison, we establish a baseline that involves encoder feature distillation without the application of mask reweighting. The findings, as detailed in Table~\ref{tab:abl_enc}, reveal that an approach solely dependent on encoder feature distillation achieves $32.4$ AP, indicating room for improvement. 
Subsequently, employing only positive queries for constructing the reweighting mask yields a slight enhancement in accuracy to $32.7$ AP, as documented in the second row of Table~\ref{tab:abl_enc}.
A significant advancement is observed when incorporating hard-negative queries with a GIoU greater than $0.5$, which elevates the distillation efficacy to $33.4$ AP, as shown in the fourth row of Table~\ref{tab:abl_enc}.
Extending the inclusion of hard-negative queries with a lower GIoU threshold to $0.0$ further improves performance, resulting in $33.7$ AP.
The application of the GIoU score for adaptive weighting fine-tunes the outcome to $33.9$ AP, underscoring the effectiveness of our proposed strategies.
Conversely, the integration of easy-negative queries, which do not align closely with ground truth, detracts from the model's accuracy, reducing it by $0.8$ AP as evidenced in the third row of Table~\ref{tab:abl_enc}. This outcome highlights the critical role of query selection and the strategic use of GIoU thresholds in optimizing encoder feature distillation for DETR models.

\subsubsection{Abaltions on the Encoder-Free Distillation} 
In this section, we investigate the impact of reducing the number of encoder layers within DEtection TRansformer (DETR) models on performance metrics such as AP and FPS.
Our approach leverages encoder distillation with the novel integration of an encoder layer adapter to address performance discrepancies arising from layer reduction. 
Specifically, we demonstrate that reducing the encoder layer count to $3$ does not significantly compromise the benefits of encoder distillation, achieving a notable increase in AP by $6.1$. 
However, eliminating encoder layers entirely results in a diminished improvement of $1.8$ AP brought by distillation, attributed to challenges in aligning encoder and ResNet-based backbone features effectively.

To address this issue, we introduced an encoder layer adapter into the training regimen, which significantly enhanced performance, raising the distillation AP from $24.6$ to $27.1$.
Additionally, when the encoder layer count is adjusted to $3$, the adapter contributes an additional improvement of $0.4$ AP, underscoring its efficacy in enhancing feature alignment.
This enhancement was particularly pronounced in the $0$ encoder-layer configuration using Conditional ResNet-18 with the AGFD alone, which achieved an AP increase of $0.7$ compared to the original Conditional ResNet-18 with $6$ encoder layers. 
Remarkably, this improvement was accompanied by a more than $1.5$-fold increase in FPS, highlighting the efficiency of our approach. 
Our findings suggest that the integration of an encoder layer adapter offers a promising avenue for optimizing DETR models in terms of both accuracy and efficiency, particularly in scenarios with reduced encoder complexity.

\begin{table}[]
\centering
\caption{Ablation study for Local Alignment Prediction Distillation between different predictions .}
\label{tab:abl_dec}
\setlength{\extrarowheight}{2pt}{
\setlength{\tabcolsep}{5mm}{
\begin{tabular}{@{}ccccc@{}}
\toprule
Setting                     & AP            & AP$\rm _S$    & AP$\rm _M$    & AP$\rm _L$    \\ \midrule
Baseline                    & 26.4          & 10.2          & 28.2          & 39.7          \\
$\mathbf{Q}^+$              & 26.7          & 9.4           & 28.9          & 41.0           \\
$\mathbf{Q}^{en}$                                            & 26.1          & 10.2          & 28.4          & 39.5          \\

$\tilde{\mathbf{Q}}_{thr=0.5}$  & 27.8          & 10.4          & 30.0          & 42.5          \\
$\tilde{\mathbf{Q}}_{thr=0.1}$ & 28.7          & 10.0          & 30.7          & 45.2          \\
$\tilde{\mathbf{Q}}_{thr=0.0}$ & \textbf{29.7} & \textbf{10.7} & \textbf{31.6} & \textbf{46.5} \\
$\mathbf{Q}^{hn}_{thr=0.0}$                                            & 29.4         & 11.4         & 31.3          & 44.9          \\
\bottomrule
\end{tabular}
}
}
\end{table}

\subsubsection{Abaltions on the LAPD}
\begin{table}[]
\centering
\caption{Training time (minutes) per epoch with different alignment strategies for prediction distillation, measured in minutes on $8$ NVIDIA Tesla V$100$ GPU of $32$ GB with a total batch size of $16$.}
\setlength{\extrarowheight}{2pt}{
\begin{tabular}{@{}ccccc@{}}
\toprule
\multirow{2}{*}[-3.5ex]{Detector} & \multicolumn{1}{l}{\multirow{2}{*}[-3.5ex]{Queries}} & \multicolumn{3}{c}{Training Time per epoch}                                                                                                                                                                      \\ \cmidrule(l){3-5} 
                          & \multicolumn{1}{l}{}                         & \begin{tabular}[c]{@{}c@{}}+Global \\ Alignment\end{tabular} & \begin{tabular}[c]{@{}c@{}}+Local\\ Alignment\\ ($thr=0.0$)\end{tabular} & \begin{tabular}[c]{@{}c@{}}+Teacher\\ Inference\end{tabular} \\ \midrule
Conditional DETR~\cite{meng2021conditional}          & 300                                          & 100                                                          & 78                                                                       & 70                                                           \\
DAB DETR~\cite{liu2022dab}                  & 300                                          & 103                                                          & 82                                                                       & 74                                                           \\
DINO~\cite{zhang2022dino}                      & 900                                          & 888                                                          & 146                                                                      & 85                                                           \\ \bottomrule
\end{tabular}
}
\label{tab:dec-time}
\end{table}

In this study, we explore the efficacy of Local Alignment Prediction Distillation (LAPD) in the decoder component of Detection Transformer models through a series of six experimental groups. 
These experiments were designed to systematically evaluate the impact of different query selections on decoder prediction distillation.
We established a baseline using Conditional DETR ResNet-$18$, trained for $12$ epochs without distillation, to assess the incremental benefits of our LAPD strategy.
Initially, we focused on distillation between only positive queries from the teacher to the student model, yielding a modest improvement of $0.3$ mAP, as indicated in the second row of Table~\ref{tab:abl_dec}.
Subsequently, we experimented with distilling easy-negative predictions (GIoU with ground truth boxes $<0$), which resulted in a $0.3$ AP decrease, underscoring the ineffectiveness of distilling low-quality predictions.
A significant enhancement was observed when we distilled between hard-negative predictions (GIoU $> 0.5$) and positive predictions for each ground truth, following bipartite matching to identify distillation pairs. This approach led to a substantial increase of $1.4$ AP. By incrementally adjusting the number of hard-negative predictions included in the distillation process, we optimized performance, achieving a peak result of $29.7$ AP when the threshold was set to $0.0$.
It should be noted that since we only select queries whose GIou with objects is bigger than zero and the average object number in one image is smaller than $8$, therefore, in extreme cases we select all the the hard-negative queries to be calculated in bipartite matching are far fewer than the total query number($300$ or $900$). Besides, solely focusing on LAPD between hard-negative predictions without including positive predictions still achieved a notable $29.4$ AP, as detailed in the final row of Table~\ref{tab:abl_dec}.
These findings underscore the nuanced role of query selection in decoder prediction distillation and highlight the potential of LAPD to enhance DETR model performance through strategic distillation approaches. This work contributes to the broader understanding of effective distillation techniques in transformer-based object detection models, offering insights into optimizing model accuracy and efficiency.
\begin{table}[]
\centering
\caption{Effectiveness of our AGFD in different encoder layer settings and influence of introducing one encoder layer as the adaptor.}
\setlength{\extrarowheight}{2pt}{
\setlength{\tabcolsep}{2.5mm}{
 \begin{tabular}{@{}cccccccc@{}}
\toprule
Enc                & Distill    & Adapter    & AP            & AP$\rm _S$    & AP$\rm _M$    & AP$\rm _L$    & FPS          \\ \midrule
\multirow{3}{*}{6} &            &            & 26.4          & 10.2          & 28.2          & 39.7          & \textbf{82}  \\
                   & \checkmark &            & 33.9          & 13.7          & 36.5          & \textbf{51.7} & \textbf{82}  \\
                   & \checkmark & \checkmark & \textbf{33.9} & \textbf{13.9} & \textbf{36.7} & 51.5          & \textbf{82}  \\ \midrule
\multirow{3}{*}{3} &            &            & 25.8          & 9.7           & 27.8          & 39.1          & \textbf{99}  \\
                   & \checkmark &            & 31.9          & 12.6          & 33.8          & \textbf{49.4} & \textbf{99}  \\
                   & \checkmark & \checkmark & \textbf{32.3} & \textbf{13.4} & \textbf{34.4} & 49.1          & \textbf{99}  \\ \midrule
\multirow{3}{*}{0} &            &            & 22.8          & 9.4           & 25.1          & 34.8          & \textbf{124} \\
                   & \checkmark &            & 24.6          & 10.3          & 26.0          & 37.0          & \textbf{124} \\
                   & \checkmark & \checkmark & \textbf{27.1} & \textbf{11.3} & \textbf{29.5} & \textbf{40.1} & \textbf{124} \\ \bottomrule
\end{tabular}%
}
}
\label{tab:abl_adapter}
\end{table}

In our further experiments, detailed in Table~\ref{tab:dec-time}, we assess the impact of Local Alignment Prediction Distillation (LAPD) on reducing the computational costs associated with distilling Detection Transformer models. These experiments specifically focused on ResNet-18 backbone student models, comparing the training time per epoch across various alignment methods against a baseline that included both student training and teacher inference times.
Our results highlight a significant computational burden imposed by global alignment used by previous methods~\cite{detrdistill, d3etr}, wherein bipartite matching between all teacher and student predictions for the DINO model with $900$ queries necessitated nearly $15$ hours per epoch. In stark contrast, employing LAPD drastically reduced the training time to approximately $2$ hours, demonstrating a reduction to less than $\frac{1}{6}$ of the time required for global alignment.
This efficiency gain was not exclusive to models with a high number of queries; even for Conditional DETR and DAB DETR models, which utilize $300$ queries, LAPD's local alignment method substantially eased the computational time burden. These findings underscore the effectiveness of the LAPD in enhancing the training efficiency for distilling DETR models, offering a scalable solution that mitigates the computational cost without compromising the final performance.
The implications of these results extend beyond mere time savings, suggesting that LAPD could play a pivotal role in advancing the scalability and practical application of DETR models across various domains. By significantly reducing the computational demands of model distilling, LAPD facilitates more sustainable and accessible deployment of advanced object detection models, potentially accelerating innovation and adoption in resource-constrained environments.

\begin{figure*}[t]
\centering
\includegraphics[width=1\linewidth]{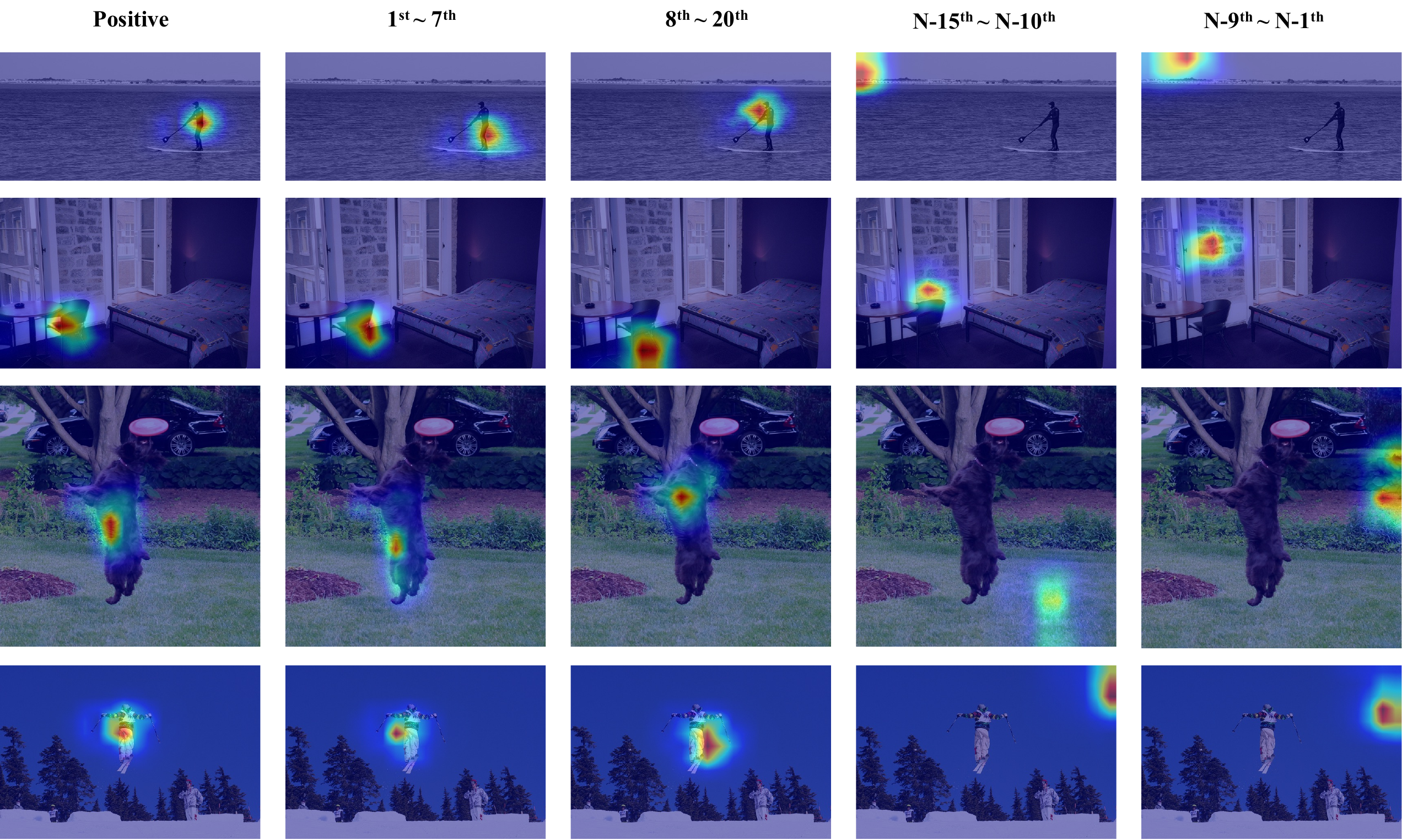}
\captionof{figure}{
Visualization of the attentive regions of different queries.
The first column features the positive prediction from bipartite matching, while the second and third columns show the randomly selected prediction from the top 20 negative predictions with the highest GIoU metrics. The last two columns present the randomly selected prediction from 15 predictions with the lowest GIoU metrics.
}
\label{fig:vis_supp}
\end{figure*}

\subsection{Visualization}

Figure \ref{fig:vis_supp} provides a nuanced visualization of how object queries correlate with respective image features, offering a selection of random examples to deepen our understanding of object perception within the model.
This figure contrasts with Figure \ref{fig:queries}, which aggregates the average features of all predictions within specific intervals in a single frame, by adopting a novel approach that involves the random selection of a single prediction from each interval for detailed visual examination. Notably, queries with higher, albeit negative, Generalized Intersection over Union (GIoU) scores demonstrate a pronounced focus on the foreground elements of objects. Conversely, queries with lower GIoU scores, which we call easy-negative queries, exhibit diminished precision in object delineation, highlighting their limited efficacy in the context of knowledge distillation.

In particular, this analysis reveals that the positive query in the first row predominantly captures the central portion of a person, omitting significant portions of the body's extremities. 
The series of randomly chosen predictions, ranging from the $1$st to the $7$th, increasingly concentrate on the lower half of the human figure.
In contrast, predictions from the $8$th to the $20$th prediction pivot towards the head region of the human. 
Similarly, the second row of Figure \ref{fig:vis_supp} demonstrates that while the positive query primarily focuses on the seat of a chair, the hard-negative queries will extend attention to encompass the chair's backrest or legs.

These insights suggest that hard-negative queries, in conjunction with positive queries, provide a more comprehensive perspective on object recognition. This nuanced understanding underscores the potential limitations of relying solely on positive queries or amalgamating all queries, including easy-negative ones, for object perception. Through this detailed visual analysis, our study contributes to the evolving discourse on enhancing knowledge distillation techniques for object detection models, advocating for a balanced inclusion of query types to achieve optimal model training and performance.

\section{Discussion and Conclusions}
\subsection{Discussion}
\label{subsec:discussion}
Initially, the cornerstone of our research is the strategic selection of queries within the context of Detection Transformer distillation. 
Recognizing the pivotal role that queries play in augmenting the distillation process, as evidenced by prior studies \cite{detrdistill, d3etr}, is fundamental to our approach. However, an observation of concern is the excessive time consumption attributed to the bipartite matching process employed for establishing distillation relationships between the student and teacher queries. This process becomes particularly burdensome when attempting to replicate experimental results in DINO, involving $900$ queries, where a single epoch of training could extend over several days, rendering the approach impractical. 
Upon further analysis of \cite{detrdistill}, it is noted that an assumption exists wherein all queries utilized in encoder distillation are perceived as less effective compared to a subset of selectively chosen positive ones, a contrast to scenarios in decoder distillation where all queries are deemed to enhance performance.
This observation raises a pivotal question regarding the specific queries that significantly contribute to the distillation process in Detection Transformers. 
Therefore, this paper proposes an innovative approach to query segmentation that goes beyond the traditional method of bipartite matching with ground truth boxes.

Moreover, it is essential to clarify our methodology's distinction from that of Group DETR \cite{chen2023group}. Unlike Group DETR, which introduces additional groups of queries to facilitate the matching of a single ground truth with multiple queries across different groups thereby expediting convergence, our approach refrains from incorporating any supplementary queries. 
Instead, we focus on categorizing the essential queries within the original query set, based solely on GIoU with ground truth, for inclusion in the precise one-to-one distillation correspondence established between student and teacher essential queries through bipartite matching.

\subsection{Conclusions and Future Work}
In this study, our Knowledge Distillation via Query Selection for Detection Transformer (QSKD) has demonstrated remarkable effectiveness across various DETR models, including Conditional DETR, DAB DETR, and DINO. Our methodology, which strategically selects informative queries to guide the distillation process via Attention-Guided Feature Distillation and Local Alignment Prediction Distillation, has shown promising results in experiments conducted on the MS-COCO dataset. For instance, we observed a significant improvement in the Conditional DETR ResNet-18 model, with AP increasing from $35.8$ to $39.9$. Compared to existing methods for the distillation of the Detection Transformer, this approach not only offers a novel perspective on utilizing queries in knowledge distillation but also contributes to enhancing model accuracy and reducing computational costs during the distillation process. 
We also investigate distillation across 
transformer modules and convolutional modules, introducing a novel approach to further reduce the computational complexity of the DETRs.

Traditional offline distillation methods, including our QSKD, depend on a pretrained teacher model to guide the student model.
To overcome this limitation, we propose to explore online distillation techniques for the DETRs in the future, eliminating the requirement to train a teacher. 
Furthermore, we plan to experiment with distillation for cross-DETR models, such as distilling Conditional DETR using DAB DETR, further verifying the applicability of our QSKD.

\bibliographystyle{elsarticle-harv} 
\bibliography{ref}
\end{document}